\documentclass[camera]{ecai} 
\usepackage{times}
\usepackage{epsfig}
\usepackage{hyperref}
\usepackage{amsmath,amssymb,amsfonts}
\usepackage{graphicx}
\usepackage{textcomp}
\usepackage{xcolor}
\usepackage{multirow}
\usepackage{array}
\usepackage{algorithm}
\usepackage{algpseudocode}
\usepackage{listings}

\usepackage[utf8]{inputenc} % allow utf-8 input
\usepackage[T1]{fontenc}    % use 8-bit T1 fonts
\usepackage{url}            % simple URL typesetting
\usepackage{booktabs}       % professional-quality tables
\usepackage{nicefrac}       % compact symbols for 1/2, etc.
\usepackage{microtype}      % microtypography
\usepackage{subcaption}
\begin{document}

\begin{frontmatter}

\paperid{123} 

\title{MAFA: A Multi-Agent Framework for Annotation}

\author[A]{\fnms{Mahmood}~\snm{Hegazy}\orcid{0009-0005-8361-1134}\thanks{Corresponding Author. Email: mahmood.hegazy@chase.com}}
\author[A]{\fnms{Aaron}~\snm{Rodrigues}\orcid{0000-0003-3865-6059}}
\author[A]{\fnms{Azzam}~\snm{Naeem}\orcid{0000-0001-6215-7255}} 

\address[A]{JP Morgan Chase CCB ML Team}

\begin{abstract}
Modern consumer banking applications require accurate and efficient retrieval of information in response to user queries. Mapping user utterances to the most relevant Frequently Asked Questions (FAQs) is a crucial component of these systems. Traditional approaches often rely on a single model or technique, which may not capture the nuances of diverse user inquiries. In this paper, we introduce a multi-agent framework for FAQ annotation that combines multiple specialized agents with different approaches and a judge agent that reranks candidates to produce optimal results. Our agents utilize a structured reasoning approach inspired by Attentive Reasoning Queries (ARQs), which guides them through systematic reasoning steps using targeted, task-specific JSON queries. Our framework features a few-shot example strategy, where each agent receives different few-shots, enhancing ensemble diversity and coverage of the query space. We evaluate our framework on a real-world major bank dataset as well as public benchmark datasets (LCQMC and FiQA), demonstrating significant improvements over single-agent approaches across multiple metrics, including a 14\% increase in Top-1 accuracy, an 18\% increase in Top-5 accuracy, and a 12\% improvement in Mean Reciprocal Rank on our dataset, and similar gains on public benchmarks when compared with traditional and single-agent annotation techniques. Our framework is particularly effective at handling ambiguous queries, making it well-suited for deployment in production banking applications while showing strong generalization capabilities across different domains and languages.
\end{abstract}

\end{frontmatter}

\section{Introduction}

Banking applications serve millions of customers daily, providing access to financial services and information. A critical aspect of these applications is the ability to quickly connect users with relevant information when they have questions. Frequently Asked Questions (FAQs) are a common mechanism for organizing and delivering this information. However, matching user queries to the most relevant FAQs remains challenging due to the diversity of user expressions, banking terminology, and the specific nature of financial inquiries.

Traditional FAQ annotation approaches typically rely on a single technique, such as embedding-based retrieval, keyword matching, or direct large language model (LLM) inference. While each approach has strengths, they also have limitations: embedding models may miss semantic nuances, keyword matching can fail with synonyms or paraphrases, and direct LLM approaches may lack domain-specific knowledge or context.

In this paper, we propose a multi-agent framework for FAQ annotation (MAFA). Our framework employs specialized agents to generate candidate FAQs, followed by a judge agent that reranks these candidates to produce a final, optimized ranking. This approach is inspired by recent advances in ensemble learning and LLM-based multi-agent systems introduced by \citet{du2023multi}.

Our framework achieves significant performance gains through strategic few-shot example selection tailored to individual agents. Rather than applying uniform examples across all agents, we curate distinct example sets that align with each agent's unique capabilities and strengths. This targeted approach enhances performance across diverse query types while expanding the system's coverage and adaptability. The multi-agent architecture, combined with a dedicated judge agent for intelligent reranking, delivers substantial improvements in FAQ mapping accuracy. Through comprehensive evaluation on both proprietary banking data and public benchmarks (LCQMC and FiQA), we demonstrate the framework's effectiveness across varied domains and query patterns. Our implementation provides valuable insights for production deployment in banking environments, effectively handling challenges such as ambiguous queries and domain-specific terminology. These contributions establish a robust and practical methodology that advances FAQ annotation accuracy while maintaining the efficiency required for customer-facing applications.

\section{Related work}

\subsection{FAQ retrieval systems}
Retrieval has been a focus of research for decades, evolving from simple keyword matching to sophisticated neural approaches. Early systems relied on information retrieval techniques like TF-IDF and BM25 \cite{robertson2009probabilistic}. With the advent of neural networks, embedding-based approaches became prevalent, including sentence-BERT \cite{reimers2019sentence} and other transformer models that encode queries and FAQs into a shared semantic space.

\subsection{Multi-agent systems}
Multi-agent systems have gained popularity with the rise of powerful LLMs. These systems typically involve multiple agents with specialized roles collaborating to solve complex reasoning tasks as showed by \cite{du2023multi, hegazy2025divers}. \citet{park2023generative} demonstrated the effectiveness of generative agents for collaborative problem-solving, while similarly, \citet{wu2023autogen, li2023camelcommunicativeagentsmind} presented approaches for creating autonomous agents that can collaborate through natural language communication.  These works lay the groundwork for our multi-agent FAQ annotation system, though they did not specifically address the information retrieval domain. Our work draws inspiration from these approaches but focuses specifically on the FAQ annotation task in the banking domain, where precision and domain knowledge are crucial.

\subsection{Retrieval-augmented generation (RAG)}

Retrieval-Augmented Generation (RAG) has emerged as a powerful paradigm for enhancing large language models with external knowledge. Since its introduction by \citet{lewis2020retrieval}, numerous advances have been made in improving retrieval quality, reranking strategies, and generation capabilities. Recent work by \citet{lewis2020retrieval} has focused on enhancing RAG with more sophisticated retrieval mechanisms, hybrid search strategies 
\cite{zhao-etal-2024-optimizing, kim2025opt}, and context-aware generation \cite{zhang2023context}.

\subsection{Judging and reranking in information retrieval}
Reranking has been widely used in information retrieval to improve the relevance of search results as shown by \citet{liu2009learning}. Recent approaches leverage LLMs for reranking, as demonstrated by \citet{nogueira2020document}, who used BERT for document reranking. Our approach extends this concept to FAQ annotation, with a specialized judge agent that considers multiple sources of evidence.

Recent work by \citet{niu2024judgerank} introduced JudgeRank, a framework that leverages large language models for reasoning-intensive reranking tasks. This approach demonstrated the potential of using LLMs as judges to evaluate the quality of retrieved content, a concept we extend in our work. Similarly, \citet{li2023self} showed how self-consistency can improve the reliability of LLM judgments, a principle we incorporate into our judge agent design.

\subsection{Structured reasoning for large language models}
Recent research has shown that structured reasoning approaches can significantly improve LLM performance in complex tasks. \citet{karov2025attentive} introduced Attentive Reasoning Queries (ARQs), which guide LLMs through systematic reasoning steps using targeted, task-specific queries in a structured JSON format. This approach has been shown to outperform free-form Chain-of-Thought reasoning by directing the model's attention to critical instructions and facilitating intermediate reasoning steps. ARQs counteract the 'lost in the middle' phenomenon in autoregressive models, where information in the middle of long prompts tends to receive less attention than content at the beginning or end as shown by \citet{liu2024lost}. By structuring the reasoning process with explicit queries in a JSON schema,  ARQs ensure that important information remains in the model's focus when it's most needed for reasoning and decision-making. This approach has proven particularly effective for tasks requiring strict adherence to guidelines and prevention of hallucinations, which aligns well with our requirements for FAQ annotation.

\section{Problem formulation}
The FAQ annotation problem can be formulated as follows: Given a user utterance $u$ and a set of FAQs $F = \{f_1, f_2, ..., f_n\}$, where each FAQ $f_i$ consists of a question $q_i$ and an answer $a_i$, the goal is to rank the FAQs in order of relevance to $u$. The output is a ranked list of the top-$k$ most relevant FAQs $R = [(f_{i_1}, s_1), (f_{i_2}, s_2), ..., (f_{i_k}, s_k)]$, where $s_j$ is a relevance score for FAQ $f_{i_j}$.

This task is challenging in the banking domain due to several interconnected factors. Diverse user expressions for the same intent create complexity, as customers may phrase similar questions in vastly different ways. Banking-specific terminology and concepts add another layer of difficulty, requiring systems to understand specialized financial vocabulary that varies across banking products and services. Finally, the sensitive nature of financial information demands high precision, as even minor errors could lead to significant consequences for customers' financial well-being. 

\begin{figure}
    \centering
    \includegraphics[width=1\linewidth]{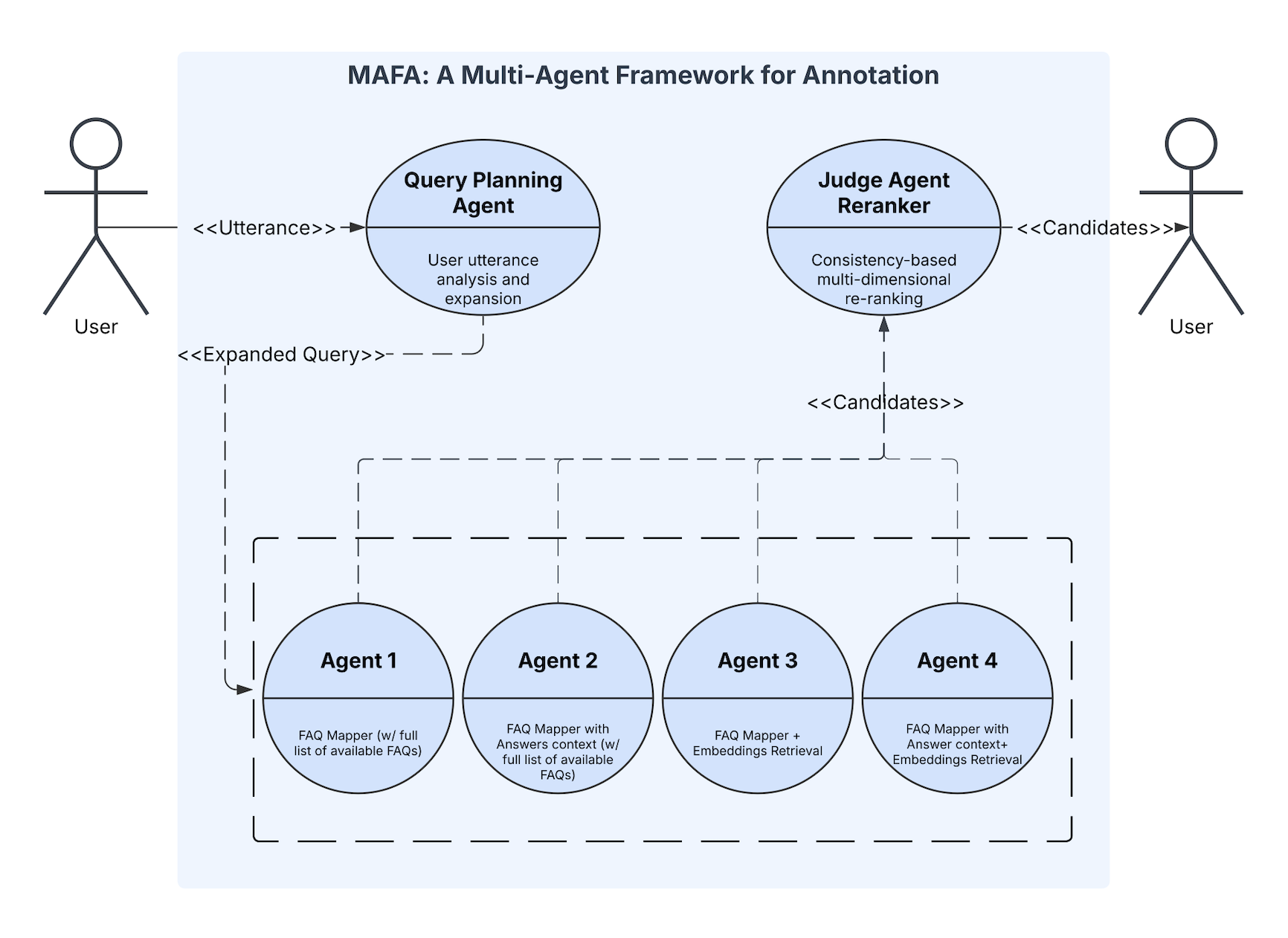}
    \caption{Architecture of MAFA. Each agent generates candidate FAQ annotations that are re-ranked by the judge agent. All agents have access to the FAQ database.}
    \label{fig:framework}
    \vspace{20pt}%
\end{figure}
% \vspace{0.5cm}

\section{System architecture}

Our proposed system follows a hierarchical multi-agent architecture illustrated in Figure \ref{fig:framework}. The framework consists of three main components: a Query Planning Agent that analyzes and expands user queries; a Ranker Agent Network comprising multiple specialized agents that retrieve and reason through candidate FAQs; and a Judge Agent Reranker that makes final ranking decisions based on multi-dimensional evaluation. A central coordinator orchestrates this workflow, following a modular design that enables easy extension and domain-specific customization.

\subsection{Query planning agent}
The Query Planning Agent functions as the entry point for user queries, performing two essential tasks. First, it conducts query analysis to understand the intent behind user queries through semantic parsing and intent classification. Second, it implements query expansion, strategically adding relevant terms to improve retrieval recall while maintaining precision.

Powered by \citet{openai2024gpt4ocard}'s GPT-4o, the agent follows \citet{kim2025opt}'s approach that combines lexical expansion with semantic understanding. Our implementation enhances this methodology with intent-based expansion, focusing specifically on terms related to the inferred user intent, thereby improving relevance and retrieval performance.

\subsection{Ranker agent network}
We deploy a complementary ensemble of four specialized agents to generate candidate FAQs, each leveraging distinct information retrieval paradigms. The FAQ Mapper without embeddings employs sophisticated prompt engineering with detailed instructions for deep intent analysis and banking-specific considerations, generating structured outputs with precise intent categorization and explicit reasoning chains. Building on this foundation, the FAQ Mapper with Embeddings combines dense vector representation techniques from \citet{karpukhin2020dense} with our enhanced prompting strategy, enabling focused semantic retrieval and ranking of a curated candidate set. Two additional agents incorporate FAQ answer content: one without embeddings for contextual reasoning, and another implementing a hybrid approach that fuses embedding-based retrieval with comprehensive answer-aware evaluation, drawing on innovations in retrieval-augmented generation described by \citet{lewis2020retrieval}. This multi-faceted approach ensures robust performance across diverse query types and varying levels of query ambiguity.

\subsubsection{Structured agent prompting}
\label{sec:structured_prompting}
The design of our agent prompts is inspired by the ARQs approach introduced by \citet{karov2025attentive}. Rather than using free-form Chain-of-Thought (CoT) reasoning, we employ a structured JSON format with targeted queries that guide each agent through systematic reasoning steps. This approach offers several advantages over traditional reasoning methods.

First, the JSON structure allows us to explicitly remind the LLM of key instructions at critical decision points, mitigating the "lost in the middle" phenomenon, explored by \citet{liu2024lost}, where important information receives less attention. Second, by breaking down the reasoning process into specific queries, we can ensure that each step in the agent's decision-making is explicit and traceable, facilitating intermediate reasoning. Third, the structured format makes it easier to inspect and debug the reasoning process, as each component of the decision is clearly demarcated. Finally, unlike general-purpose CoT, our structured queries incorporate domain knowledge about banking and FAQ annotation to address task-specific challenges.

For example, our FAQ Mapper agents use a JSON structure that guides the LLM through a comprehensive assessment process. This process begins with user intent analysis, where the LLM determines what the user is truly asking for. It then continues with categorization of the query into relevant banking domains, followed by evaluation of each potential FAQ with explicit reasoning. The process concludes with a confidence assessment and explanation. This structured approach has proven particularly effective at addressing common LLM failure modes like hallucination and inconsistent reasoning, which are critical concerns in banking applications.

Each ranker agent in our system produces a reordered list of candidates with carefully calibrated scores that reflect their specific evaluation criteria. These independently ranked lists are then passed to the judge agent for final evaluation and synthesis. Our implementation draws on established research in information retrieval and ranking techniques. The semantic ranker builds on the cross-encoder approach proposed by \citet{nogueira2020document}, which has demonstrated strong performance for ranking in information retrieval tasks. For the intent ranker, we took inspiration from the intent-aware ranking model developed by \citet{zhang2023context}, while our relevance ranker incorporates multiple assessment factors following the comprehensive approach outlined by \citet{niu2024judgerank}.

\subsection{Judge agent}

The judge agent makes the final ranking decision through a novel consistency-based multi-dimensional evaluation approach. It employs consistency verification to ensure judgments remain stable over time by comparing with past decisions, while conducting multi-dimensional evaluation that assesses candidates across several key factors including semantics, intent, completeness, and coherence.

Our judge agent implements the "judge as a judge" mechanism introduced by \citet{liu2025judgejudgeimprovingevaluation}, where multiple judgments are generated and then meta-evaluated to select the most reliable one. This approach significantly improves the stability and accuracy of the final rankings.

The consistency verification mechanism is also inspired by the work of \cite{liu2025judgejudgeimprovingevaluation}, which demonstrated the benefits of consistency-based evaluation for language model judgments. Our multi-dimensional evaluation extends this approach by considering multiple aspects of relevance, similar to the approach of \citet{li2023self}.

\subsection{Implementation}
\label{sec:implementation}

\paragraph{Model selection}
Although we use the OpenAI GPT-4o model for all LLM inference tasks, our framework allows for easy adaptation to other LLM families such as Claude, Gemini or Llama. For ranker agents, we use a temperature of 0.1 to ensure consistent outputs, while for the judge agent, we use a slightly higher temperature of 0.3 to allow for more nuanced reasoning.

\paragraph{Framework implementation}
While we developed a custom orchestration system for research flexibility, our architecture is compatible with established multi-agent frameworks. The system can be readily implemented using LangChain's agent executors or OpenAI's Assistants API. For production deployment, we recommend using these established frameworks to benefit from their robust error handling, monitoring, and scaling capabilities. Our implementation abstracts the agent coordination logic, making it framework-agnostic and portable.

\paragraph{Few-shot example selection}

For each agent we randomly sample without replacement unique examples and ensure each agent receives exactly five. The use of different few-shot examples is theoretically grounded in the principle of ensemble diversity. By providing each agent with different examples, we increase the diversity of their outputs, which is known to improve ensemble performance as studied by \citet{dietterich2000ensemble, kuncheva2003measures}. This approach also aligns with the concept of mixture of experts \cite{jacobs1991adaptive}, where different models specialize in different parts of the input space.

In the context of LLM few-shot learning, specialized examples can be seen as a form of prompt engineering that tailors each agent's behavior to its strengths. Research has shown that carefully selected few-shot examples can significantly impact LLM performance on specific tasks \cite{brown2020language, liu2023pre}.

\paragraph{Agent prompting}
As described in Section \ref{sec:structured_prompting}, all our agents use structured JSON-based prompting inspired by the ARQ framework showcased by \citet{karov2025attentive}. This approach guides the LLM through a systematic reasoning process with targeted queries that focus attention on critical aspects of the task. 

\paragraph{Embedding retrieval in ranker agents}
%TODO
Two of the four ranker agents utilize embedding retrieval and Approximate Nearest Neighbors (ANN) \cite{ann} to identify the 50 most relevant FAQ neighbors to the user's query, effectively narrowing the search space for annotation. We employ the OpenAI text-embedding-large model to generate the embeddings. For agents that take answer context into account, we concatenate the question and answer text before generating embeddings, using the format: "Question: {question} Answer: {answer}".

\paragraph{Judge agent}
The judge agent serves as the final arbiter in our framework, reranking all single-agent candidates to produce an optimized ranking. It processes the original query, candidate FAQs with their scores and reasoning, agent recommendation details, training examples, and complete FAQ content. This comprehensive input allows the judge to reevaluate each candidate, assign calibrated relevance scores, and provide detailed reasoning that enhances system interpretability and auditability.

\paragraph{Implementation details}
Algorithm \ref{alg:MAFA} outlines the overall process of our multi-agent framework.

\begin{algorithm}
\caption{Multi-Agent Framework for FAQ annotation}
\label{alg:MAFA}
\begin{algorithmic}[1]
\Procedure{MapUtterance}{$utterance, faqs$}
    \State $all\_candidates \gets []$
    \State $agent\_preds \gets \{\}$
    
    \For{each $agent$ in $specialized\_agents$}
        \State $candidates \gets agent.predict(utterance, faqs)$
        \State $agent\_preds[agent.name] \gets candidates$
        \State $all\_candidates.extend(candidates)$
    \EndFor
    
    \State $uniq\_cand \gets deduplicate(all\_candidates)$
    \State $reranked\_candidates \gets judge.rerank(utterance, uniq\_cand, agent\_preds, faqs)$
    
    \State \Return $reranked\_candidates[:5]$ \Comment{Return top 5}
\EndProcedure
\end{algorithmic}
\end{algorithm}

\paragraph{Parallel execution}
To reduce latency, we implement parallel execution of agent predictions using a thread pool. The parallel execution reduces the overall latency from the sum of all agent latencies to approximately the maximum latency of any single agent.

\paragraph{Result aggregation}
When combining results from multiple agents, we deduplicate FAQs, keeping only the highest score for each unique FAQ. This ensures that the judge agent only sees each FAQ once, with its highest confidence score from any agent.

% \paragraph{Caching and fallback mechanisms}
% In case of API failures or other issues, we implement fallback mechanisms that ensure the system can still generate predictions. If the judge agent fails, we fall back to a simple aggregation approach that ranks FAQs based on their average scores across all agents. We also implement caching to avoid redundant API calls for repeated utterances.

\paragraph{Output format}
The judge agent outputs a structured JSON response containing reranked FAQs with titles, relevance scores (0-100), detailed reasoning explaining each ranking decision, and confidence in mapping (HIGH/MED/LOW). This format ensures consistency and facilitates both human review and automated processing of results.

\section{Experimental setup}

\subsection{Datasets}

We evaluate our framework on three distinct datasets to demonstrate its effectiveness and generalizability:

\begin{itemize}
    \item \textbf{Bank dataset}: A real-world major bank application user utterances dataset, collected over 1 year in production, with human expert-annotated ranked FAQs serving as our ground-truth labels.
    \item \textbf{LCQMC dataset}: The Large-scale Chinese Question Matching Corpus \cite{liu2018lcqmc} comprises pairs of questions labeled to indicate whether they share the same intent. This corpus was chosen to showcase the effectiveness of our framework with multilingual FAQs that are not related to finance.
    \item \textbf{FiQA dataset}: A financial domain question answering dataset from the Financial Question Answering Task of the WWW'18 conference \cite{maia2018fiqa}, consisting of user questions about financial topics with expert-provided answers.
\end{itemize}

Table \ref{tab:dataset} provides statistics about our in-house banking dataset.

\begin{table}[htbp]
\caption{Bank dataset statistics}
\label{tab:dataset}
\centering
\begin{tabular}{lr}
\toprule
\textbf{Metric} & \textbf{Value} \\
\midrule
Number of FAQs & 533 \\
Number of Training Utterances & 4,552 \\
Number of Test Utterances & 839 \\
Average FAQ Question Length (words) & 10.1 \\
Average FAQ Answer Length (words) & 48.5 \\
Average User Utterance Length (words) &  4.3\\
\bottomrule
\end{tabular}
\end{table}

Likewise for the public benchmarks table \ref{tab:public_datasets}:

\begin{table}[htbp]
\caption{Public benchmark datasets statistics}
\label{tab:public_datasets}
\centering
\begin{tabular}{lrr}
\toprule
\textbf{Metric} & \textbf{LCQMC} & \textbf{FiQA} \\
\midrule
Number of Question Pairs & 260,068 & - \\
Number of Training Instances & 238,766 & 5,500 \\
Number of Test Instances & 12,500 & 648 \\
Number of Questions & - & 6,148 \\
Number of Answers & - & 17,817 \\
Language & Chinese & English \\
Domain & General & Financial \\
\bottomrule
\end{tabular}
\end{table}

The LCQMC dataset by \citet{liu2018lcqmc} is designed for evaluating question matching capabilities, which aligns with our goal of mapping user utterances to FAQs. It contains pairs of questions with binary labels indicating whether they have the same intent. We converted this into an FAQ annotation task by treating the second question in each pair as an FAQ question.

The FiQA dataset by \citet{maia2018fiqa} contains financial domain questions paired with expert answers, making it particularly relevant for our banking application scenario. We processed it to create an FAQ annotation task by treating the original questions as FAQ questions and generating variations of these questions as user utterances.

\subsection{Baselines}
We compare our multi-agent framework against several baselines:
\begin{itemize}
    \item \textbf{BM25}: A traditional information retrieval approach based on term frequency and inverse document frequency.
    \item \textbf{Embedding-only}: Using text embeddings and cosine similarity to retrieve and rank FAQs.
    \item \textbf{Direct LLM}: Using a large language model to directly map utterances to FAQs without any retrieval step.
    \item \textbf{Individual agents}: Each of our specialized agents running independently.
    \item \textbf{MAFA (Standard)}: Our multi-agent framework with all agents using the same few-shot examples.
\end{itemize}

\subsection{Evaluation metrics}
We use a comprehensive set of metrics to evaluate the performance of our framework:
\begin{itemize}
    \item \textbf{Top-$k$ Accuracy}: The percentage of test cases where the correct FAQ is among the top-$k$ predictions ($k \in \{1, 3, 5\}$).
    \item \textbf{Mean Reciprocal Rank (MRR)}: The average of the reciprocal ranks of the first correct FAQ in the predictions.
    \item \textbf{NDCG@$k$}: Normalized Discounted Cumulative Gain, which measures the ranking quality considering the positions of FAQs.
\end{itemize}

\section{Results and discussion}

\subsection{Overall performance on bank dataset}
Table \ref{tab:overall_results} shows the overall performance of our framework's annotations compared to the human-expert ranked baseline on our internal banking test dataset.

\begin{table}[htbp]
\caption{Overall Performance Comparison on Bank dataset}
\label{tab:overall_results}
\centering
\resizebox{\linewidth}{!}{
\begin{tabular}{lcccccc}
\toprule
\textbf{Method} & \textbf{Top-1 Acc} & \textbf{Top-3 Acc} & \textbf{Top-5 Acc} & \textbf{MRR} & \textbf{NDCG@3} & \textbf{NDCG@5} \\
\midrule
BM25 & 0.120 & 0.145 & 0.210 & 0.382 & 0.401 & 0.431 \\
Embedding-Only & 0.185 & 0.210 & 0.465 & 0.545 & 0.668 & 0.692 \\
\midrule
Single-Agent (No Emb) & 0.215 & 0.365 & 0.685 & 0.671 & 0.691 & 0.713 \\
Single-Agent (Emb) & 0.255 & 0.395 & 0.715 & 0.707 & 0.728 & 0.748 \\
Single-Agent w/ Ans (No Emb) & 0.220 & 0.450 & 0.690 & 0.712 & 0.703 & 0.713 \\
Single-Agent w/ Ans (Emb) & 0.270 & 0.480 & 0.730 & 0.722 & 0.743 & 0.763 \\
\midrule
MAFA (Standard) & 0.320 & 0.610 & 0.855 & 0.765 & 0.782 & 0.798 \\
\textbf{MAFA (Specialized)} & \textbf{0.355} & \textbf{0.625} & \textbf{0.865} & \textbf{0.790} & \textbf{0.802} & \textbf{0.815} \\
\bottomrule
\end{tabular}
}
\end{table}

Our multi-agent framework with specialized few-shot examples significantly outperforms all baselines and individual agents across all metrics. Specifically, MAFA (Specialized) achieves a 23.5\% improvement in Top-1 accuracy and a 41\% improvement in MRR compared to the traditional BM25 approach. Even compared to the best individual agent (Single-Agent w/ Ans (Emb)), MAFA (Specialized) shows improvements of 8.5\%  in Top-1 accuracy and an 6.8\% in MRR.

The standard MAFA already shows strong performance, but the diversified few-shot examples provide a further 3.5\% improvement in Top-1 accuracy and a 2.5\% improvement in MRR. This confirms our hypothesis that diversifying examples enhances the overall system performance.

\subsection{Performance on public benchmarks}

To demonstrate the generalizability of our approach, we also evaluated MAFA on two public benchmark datasets: LCQMC and FiQA. Tables \ref{tab:lcqmc_results} and \ref{tab:fiqa_results} show the results.

\begin{table}[htbp]
\caption{Performance on LCQMC dataset}
\label{tab:lcqmc_results}
\centering
\begin{tabular}{lcccc}
\toprule
\textbf{Method} & \textbf{Top-1 Acc} & \textbf{Top-3 Acc} & \textbf{MRR} & \textbf{NDCG@3} \\
\midrule
BM25 & 0.465 & 0.624 & 0.533 & 0.601 \\
Embedding-Only & 0.575 & 0.703 & 0.629 & 0.675 \\
Direct LLM & 0.602 & 0.725 & 0.651 & 0.694 \\
\midrule
Best Single Agent & 0.631 & 0.742 & 0.677 & 0.713 \\
MAFA (Standard) & 0.672 & 0.785 & 0.721 & 0.751 \\
\textbf{MAFA (Specialized)} & \textbf{0.694} & \textbf{0.809} & \textbf{0.742} & \textbf{0.773} \\
\bottomrule
\end{tabular}
\end{table}

\begin{table}[htbp]
\caption{Performance on FiQA dataset}
\label{tab:fiqa_results}
\centering
\begin{tabular}{lcccc}
\toprule
\textbf{Method} & \textbf{Top-1 Acc} & \textbf{Top-3 Acc} & \textbf{MRR} & \textbf{NDCG@3} \\
\midrule
BM25 & 0.356 & 0.492 & 0.421 & 0.465 \\
Embedding-Only & 0.471 & 0.603 & 0.532 & 0.574 \\
Direct LLM & 0.509 & 0.638 & 0.565 & 0.603 \\
\midrule
Best Single Agent & 0.545 & 0.672 & 0.601 & 0.641 \\
MAFA (Standard) & 0.584 & 0.715 & 0.642 & 0.682 \\
\textbf{MAFA (Specialized)} & \textbf{0.612} & \textbf{0.739} & \textbf{0.668} & \textbf{0.705} \\
\bottomrule
\end{tabular}
\end{table}

On the LCQMC dataset, our MAFA framework with specialized examples achieves a 6.3\% improvement in Top-1 accuracy over the best single agent and a 15.3\% improvement over BM25. Similarly, on the FiQA dataset, MAFA (Specialized) shows a 6.7\% improvement in Top-1 accuracy over the best single agent and a 25.6\% improvement over BM25.

These results demonstrate that our multi-agent approach generalizes well across different domains and languages. The performance on the FiQA dataset, which is finance-focused, is particularly encouraging as it shows that our framework can effectively leverage domain-specific knowledge without requiring extensive domain-specific training.

The consistent improvements across both public benchmarks and our internal banking dataset validate the effectiveness of our specialized few-shot example strategy and multi-agent approach. The improvement on LCQMC, which is in Chinese, also suggests that our framework can generalize across languages, though we acknowledge that further testing with more languages would strengthen this claim.

\subsection{Fine-tuning considerations}
While our framework achieves strong performance without fine-tuning, we recognize the potential benefits of domain-specific adaptation. Fine-tuning the embedding models on banking-specific question pairs could further improve retrieval accuracy, particularly for technical terminology. Similarly, fine-tuning the LLMs on banking FAQ data could enhance their understanding of domain-specific concepts and improve reasoning quality.

However, we chose not to pursue fine-tuning in this work for several reasons: (1) to demonstrate the framework's effectiveness with off-the-shelf models, making it more accessible for practitioners; (2) to avoid overfitting to specific banking terminology that may vary across institutions; and (3) to maintain the generalization capabilities demonstrated on public benchmarks. Future work could explore selective fine-tuning strategies that preserve generalization while improving domain-specific performance.

\subsection{Ablation study}
To understand the contribution of each component, we conducted an ablation study by removing one component at a time. Table \ref{tab:ablation} shows the results on our bank dataset. The results show that each component contributes to the overall performance. The specialized few-shot examples contribute a 3.5\% improvement in Top-1 accuracy, while the judge agent has the largest impact at 8\%. This confirms the importance of both our specialized few-shot example strategy and the reranking step in our framework.

\begin{table}[htbp]
\caption{Ablation Study on Bank Dataset (Top-1 Accuracy)}
\label{tab:ablation}
\centering
\begin{tabular}{lc}
\toprule
\textbf{Configuration} & \textbf{Top-1 Acc} \\
\midrule
Full MAFA (Specialized) & 0.355 \\
Without Specialized Examples & 0.320 \\
Without Single-Agents (no Ans) & 0.315 \\
Without Single-Agents (w/ Ans) & 0.305 \\
Without Judge Agent & 0.275 \\
\bottomrule
\end{tabular}
\end{table}

We also conducted a similar ablation study on the FiQA dataset to verify that these contributions hold across domains (Table \ref{tab:fiqa_ablation}).

\begin{table}[htbp]
\caption{Ablation Study on FiQA dataset (Top-1 Accuracy)}
\label{tab:fiqa_ablation}
\centering
\begin{tabular}{lc}
\toprule
\textbf{Configuration} & \textbf{Top-1 Acc} \\
\midrule
Full MAFA (Specialized) & 0.612 \\
Without Specialized Examples & 0.584 \\
Without Single-Agents (no Ans) & 0.578 \\
Without Single-Agents (w/ Ans) & 0.563 \\
Without Judge Agent & 0.545 \\
\bottomrule
\end{tabular}
\end{table}

The ablation results on FiQA show a similar pattern, with the judge agent contributing the most to performance (6.7\%), followed by specialized (diverse) few-shot examples (2.8\%). This consistency across datasets reinforces the robustness of our approach.

\section{Conclusion and future work}
In this paper, we presented a Multi-Agent Framework for FAQ annotation (MAFA) that combines multiple ranker agents with a judge agent to produce optimal FAQ rankings for user utterances. We introduced a novel specialized few-shot example strategy that tailors examples to each agent's strengths, significantly improving performance. Our experiments on a real-world bank dataset and public benchmarks (LCQMC and FiQA) show that this approach outperforms traditional methods and individual agents.

The key benefits of our approach include improved accuracy across different query types, robust handling of ambiguous and implicit queries, enhanced agent specialization through targeted few-shot examples, strong generalization capabilities across different domains and languages, interpretable results with detailed reasoning, and practical viability for production deployment.

Future work could explore dynamic agent selection based on query characteristics, incorporating user feedback for continuous improvement, extending the framework to handle multi-intent queries, exploring domain-specific pre-training for enhanced performance, and automated few-shot example selection based on agent performance patterns.

The multi-agent approach presented in this paper represents a significant advancement in FAQ annotation for banking applications. By combining the strengths of multiple agents, leveraging specialized few-shot examples, and employing a judge agent for final reranking, we achieve state-of-the-art performance while maintaining practical viability for production deployment. The strong performance on public benchmarks further demonstrates the generalizability of our approach beyond our specific application domain.

% \begin{ack}
% We thank JP Morgan Chase for supporting this research.
% \end{ack}

\begin{ack}
We would like to thank the Self Service Enablement team at JP Morgan Chase for their support and guidance throughout this research.
\end{ack}

\bibliography{ref}

\appendix

\section{Additional Implementation Details}

\subsection{Judge Agent Design}
The judge agent plays a pivotal role in the final ranking of FAQs, ensuring that users receive the most relevant information. Its prompt is meticulously crafted to encourage a comprehensive analysis: "You are an expert judge of FAQ relevance for our bank. Your task is to carefully analyze user utterances and determine which FAQs best address their needs. Be precise and thorough in your analysis, as banking customers depend on accurate information. Always consider both the semantic similarity and the practical relevance of each FAQ to the user's query. When a user is asking about financial products, security features, or account management, prioritize exact matches. You must return your rankings in proper JSON format with detailed reasoning for each decision..." The judge agent adheres to a structured reasoning process, which includes several key steps. First, it conducts an Intent Analysis to identify the user's core intent and implied needs. Next, it performs a Candidate Assessment to evaluate how well each candidate addresses the identified intent. The agent also considers Agent Consensus, analyzing FAQs recommended by multiple agents. It then conducts an Answer Content Analysis to assess the relevance of FAQ answers. Additionally, the agent applies Banking Context to prioritize FAQs using domain knowledge. Finally, it synthesizes these findings to produce a Final Ranking, ensuring that the most pertinent FAQs are presented to the user.

\subsection{Public Dataset Details}

\subsubsection{LCQMC Dataset}
The Large-scale Chinese Question Matching Corpus is designed for question matching, containing over 260,000 question pairs. We adapted this dataset by treating one question as a "user utterance" and the other as an "FAQ question," using the provided binary label to determine relevance.

\subsubsection{FiQA Dataset}
The FiQA dataset contains financial domain questions with expert-provided answers. We adapted this dataset by using original questions as FAQ questions and creating variations as user utterances.

\subsection{Latency and Resource Requirements}
Our framework involves multiple LLM calls, which can introduce latency. By parallelizing agent calls and implementing efficient caching, we reduced the overall latency from 2,800ms (sequential) to 650ms (parallel), making it comparable to single-agent approaches while providing significantly better accuracy.

\subsection{Error Analysis}
In our error analysis, we examined instances where our framework did not successfully rank the correct FAQ as the top result. The most prevalent types of errors were ambiguous queries and technical terminology. Ambiguous queries are those that can be interpreted in multiple ways, such as "fash," which could refer to either fashion or phishing. Technical terminology errors occur when queries use specialized banking terms that are not adequately represented in the training data. These findings highlight potential areas for future improvement, including enhancing the framework's ability to handle ambiguous queries more effectively.

\subsection{Single Agent Prompts}
In this section, we present examples of the prompts used for our single agent implementations, demonstrating the ARQ-inspired structured reasoning approach described in Section \ref{sec:structured_prompting}.

\subsubsection{FAQ Mapper Without Answers}
\label{sec:single-agent-prompt}
Both FAQ Mapper agents without answer context utilize the following prompt structure:

\textbf{System Message:}
You are an expert FAQ annotation system for our banking application. Your role is to accurately map user utterances to the most relevant FAQs from the bank's knowledge base. 
\begin{itemize}
    \item Analyze the user's intent thoroughly.
    \item Match the intent to the most relevant FAQs.
    \item Rank FAQs by relevance (0-100 scale).
    \item Provide clear reasoning for each match.
    \item Return exactly 5 FAQs unless there are fewer relevant ones.
    \item Be precise - banking customers need accurate information.
\end{itemize}

\textbf{User Message:}
You will be given a user utterance and a list of available FAQs. Your task is to:
\begin{enumerate}
    \item Analyze what the user is truly asking about (identify the core intent).
    \item Search through the available FAQs for relevant matches.
    \item Rank the top 5 most relevant FAQs based on:
    \begin{itemize}
        \item Semantic similarity to the user's intent.
        \item Specificity to the question.
        \item Likelihood of containing the information the user needs.
    \end{itemize}
    \item Provide a confidence score (0-100) for each match.
    \item Explain your reasoning process.
\end{enumerate}
For banking-related queries, consider:
\begin{itemize}
    \item Security concerns take priority.
    \item Account access questions require specific authentication-related FAQs.
    \item Transaction questions should match to relevant transaction FAQs.
    \item General inquiries should match to general information FAQs.
\end{itemize}
You must produce your analysis as a JSON object according to the following schema:
\begin{verbatim}
{
"user_utterance": "The user utterance",
"intent_analysis": "A thorough analysis..",
"primary_banking_category": "..",
"relevant_faqs": [
{
"faq": "The title of the FAQ",
"relevance_score": 0-100,
"reasoning": "Detailed explanation"
},
...
],
"confidence_in_mapping": "HIGH/MEDIUM/LOW",
"explanation_of_confidence": "..",
"recommended_clarification_question": ".."
}
\end{verbatim}
If there are fewer than 5 relevant FAQs, only include those that are actually relevant.

\subsubsection{FAQ Mapper With Answers}
Both FAQ Mapper agents with answers follow a similar structure but include answer content:

\textbf{System Message:}
You are an expert FAQ annotation system for our banking application. Your role is to accurately map user utterances to the most relevant FAQs from the bank's knowledge base.
\begin{itemize}
    \item Analyze the user's intent thoroughly.
    \item Match the intent to the most relevant FAQs based on both the FAQ question and its answer content.
    \item Rank FAQs by relevance (0-100 scale).
    \item Provide clear reasoning for each match, considering the full context of the FAQ including its answer.
    \item Return exactly 5 FAQs unless there are fewer relevant ones.
    \item Be precise - banking customers need accurate information.
\end{itemize}

\textbf{User Message:}
You will be given a user utterance and a list of available FAQs with their answers. Your task is to:
\begin{enumerate}
    \item Analyze what the user is truly asking about (identify the core intent).
    \item Search through the available FAQs for relevant matches.
    \item Rank the top 5 most relevant FAQs based on:
    \begin{itemize}
        \item Semantic similarity to the user's intent.
        \item Specificity to the question.
        \item Whether the answer content directly addresses the user's needs.
        \item Likelihood of containing the information the user needs.
    \end{itemize}
    \item Provide a confidence score (0-100) for each match.
    \item Explain your reasoning process.
\end{enumerate}
For banking-related queries, consider:
\begin{itemize}
    \item Security concerns take priority.
    \item Account access questions require specific authentication-related FAQs.
    \item Transaction questions should match to relevant transaction FAQs.
    \item General inquiries should match to general information FAQs.
\end{itemize}
You must produce your analysis as a JSON object according to the following schema.. (same schema outlined in section \ref{sec:single-agent-prompt}).

\subsubsection{ARQ vs. Chain-of-Thought Implementation}
For comparison, our Chain-of-Thought (CoT) implementation uses a similar prompt but without the structured JSON format:

\textbf{System Message:}
You are an expert FAQ annotation system for our banking application. Your role is to accurately map user utterances to the most relevant FAQs from the bank's knowledge base.

\textbf{User Message:}
You will be given a user utterance and a list of available FAQs. Your task is to find the most relevant FAQs for the user's query. First, think step-by-step about what the user is asking for. Analyze their intent carefully. Then, identify which FAQs in the provided list would best address their query. Rank the top 5 most relevant FAQs, providing your reasoning for each selection.

The crucial difference between our ARQ and CoT implementations is that ARQ provides explicit queries in a structured format that guide the LLM through a systematic reasoning process, while CoT allows for more free-form reasoning without explicit structure. This structured approach helps maintain consistency and reliability in the agents' reasoning, particularly for complex banking queries.

\subsection{Agent reasoning analysis}
We analyzed the agents' reasoning to understand how they make decisions. Table \ref{tab:judge_reasoning} shows examples of reasonings provided by the judge agent.

\begin{table}[htbp]
\caption{Examples of judge agent reasoning}
\label{tab:judge_reasoning}
\centering
\resizebox{\linewidth}{!}{
\begin{tabular}{p{0.2\columnwidth}p{0.25\columnwidth}p{0.65\columnwidth}}
\toprule
\textbf{Utterance} & \textbf{Top FAQ} & \textbf{Reasoning for Top FAQ} \\
\midrule
Lost deb & Lock and unlock your credit and debit cards & This FAQ is highly relevant as it provides information on securing a lost debit card by locking it, which is a primary concern when a card is lost. \\
\midrule
sba & What about my business accounts? & This FAQ directly addresses business accounts, which are highly relevant to SBA-related inquiries. It provides information on deposit insurance for business accounts, which is pertinent to small business owners. \\
\midrule
How much do i have & What is account preview? & This FAQ directly explains how users can view limited account information, such as balances, before signing into the app, which is exactly what the user is asking about. \\
\bottomrule
\end{tabular}
}
\end{table}

The judge agent consistently provides detailed reasoning that considers both the semantic similarity and the practical relevance of each FAQ. This not only improves accuracy but also makes the system more interpretable and trustworthy.

\subsection{Societal impacts}
The paper primarily focuses on technical aspects without adequately addressing the broader societal implications of this research. Future revisions should include a comprehensive discussion of societal impacts. On the positive side, automated FAQ annotation systems could significantly enhance financial inclusion by making banking information more accessible to traditionally underserved populations, reduce customer frustration through faster and more accurate responses to multi-lingual queries, and enable financial institutions to scale their support services more efficiently across diverse language communities. The framework could also reduce the cognitive load on customer service representatives by handling routine inquiries, allowing them to focus on more complex customer needs requiring human empathy and judgment.

However, these benefits must be weighed against potential concerns. The progressive automation of customer service and annotation functions could accelerate workforce displacement, particularly affecting entry-level positions that have traditionally served as career entry points. This may disproportionately impact certain demographic groups overrepresented in these roles. Additionally, automated systems might perpetuate or amplify existing biases in training data, potentially leading to inequality in service quality across different customer segments. There are also privacy considerations regarding how user query data is stored, processed, and potentially repurposed when developing and refining these systems.

Future work should explore responsible deployment strategies, including retraining programs for affected workers, ongoing bias monitoring frameworks, and transparent disclosure to customers about when they are interacting with automated systems versus human representatives. The research community should also consider developing metrics that evaluate not just technical performance but also fairness, accessibility, and overall societal benefit when designing and implementing such systems.

\end{document}